\pdfoutput=1

\documentclass[11pt]{article}

\usepackage[]{acl} 

\usepackage{times}
\usepackage{latexsym}

\usepackage[T1]{fontenc}

\usepackage[utf8]{inputenc}

\usepackage{microtype}
\usepackage{tikz-dependency}
\usepackage{microtype}
\usepackage{graphicx}
\usepackage{booktabs}
\usepackage{multirow}
\usepackage{xcolor,pifont}
\usepackage{float}

\definecolor{myorange}{RGB}{255,194,10}
\definecolor{myblue}{RGB}{12,123,220}
\definecolor{mygreen}{RGB}{132,220,95}
\definecolor{myred}{RGB}{213,94,0}
\definecolor{mygray}{RGB}{145,145,145}

\newcommand*\colourcheck[1]{%
  \expandafter\newcommand\csname #1check\endcsname{\textcolor{#1}{\ding{52}}}%
}
\colourcheck{blue}
\newcommand*\colourcross[1]{%
  \expandafter\newcommand\csname #1cross\endcsname{\textcolor{#1}{\ding{55}}}%
}
\colourcross{red}

\title{Improving Zero-Shot Cross-lingual Transfer Between Closely Related Languages by Injecting Character-Level Noise}

\author{Noëmi Aepli$^1$ \and Rico Sennrich$^{1,2}$\\
  $^1$Department of Computational Linguistics, University of Zurich\\
  $^2$School of Informatics, University of Edinburgh \\ \medskip
  \texttt{\{naepli,sennrich\}@cl.uzh.ch}}

\begin{document}
\maketitle
\begin{abstract}
Cross-lingual transfer between a high-resource language and its dialects or closely related language varieties should be facilitated by their similarity. However, current approaches that operate in the embedding space do not take surface similarity into account.
This work presents a simple yet effective strategy to improve cross-lingual transfer between closely related varieties. We propose to augment the data of the high-resource source language with character-level noise to make the model more robust towards spelling variations.
Our strategy shows consistent improvements over several languages and tasks: Zero-shot transfer of POS tagging and topic identification between language varieties from the Finnic, West and North Germanic, and Western Romance language branches.
Our work provides evidence for the usefulness of simple surface-level noise in improving transfer between language varieties.

\end{abstract}

\section{Introduction}

\begin{figure}
    \centering
    \includegraphics[width=0.48\textwidth]{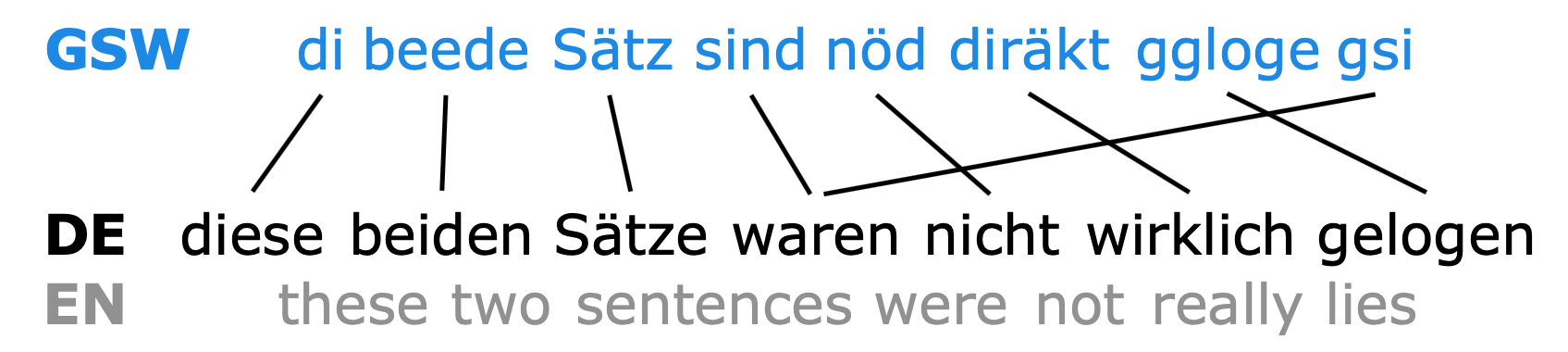}
    \caption{\textcolor{myblue}{Swiss German (GSW)} sentence with corresponding standard German (DE) and \textcolor{mygray}{English (EN)} translations. The sentence shows various spelling differences on the word level, and reordering occurs on the sentence level due to different past-tense formation.}
    \label{fig:gswexample}
\end{figure}

Recent research has achieved impressive results in zero-shot cross-lingual transfer based on multilingual pre-training~\citep{devlin-etal-2019-bert,Conneau19neurips} or monolingual transfer of embeddings~\citep{artetxe2020cross}.
However, these methods require large amounts of unlabeled data in the target language~\cite{lauscher-etal-2020-zero} and do not take into account surface similarity between languages except for the sharing of subword units in multilingual models. For the transfer between closely related languages and dialects, we deem it desirable to exploit the similarity of surface representations. Specifically, we target orthographic variations that commonly result from pronunciation differences between closely related languages. \footnote{Note that there are also differences on different levels as described in~\citet{hollenstein-aepli-2014-compilation} and partly observable in Figure \ref{fig:gswexample} which illustrates a German example sentence with a closely related variant.}

In this paper, we propose to augment the training data of a high-resource language with character-level noise to simulate spelling variations and thus facilitate generalization to closely related\footnote{Language relatedness is on a continuum, and the difference between dialects and distinct languages is often political. Hence we use a broader term to indicate that the method is not limited to dialects.} low-resource languages. 

We test this strategy on two tasks and several language regions. The considered tasks are part-of-speech (POS) tagging on the word level and topic classification on the sentence level; the languages are from the Finnic, West and North Germanic, and Western Romance language branches.
We observe that our baseline method for cross-lingual transfer learns undesirable heuristics, e.g., assigning unseen words to open word classes in POS tagging and that injecting noise reduces this bias.
Our experiments show absolute accuracy improvements between $1.4$ and $22$ percentage points over the state of the art, providing evidence that a simple data-augmentation strategy can boost transfer learning for language varieties and dialects with a closely related high-resource language.

\section{Related Work}

\paragraph{Zero-shot cross-lingual transfer} based on multilingual language models~\cite{devlin-etal-2019-bert,Conneau19neurips} or machine translation models~\cite{mmte} has turned out to be surprisingly effective.
Such representations proved themselves beneficial for a range of diverse tasks~\cite{xtreme}. However, they still require large-scale data sets to train, making them impractical for low-resource languages, to which dialects and language varieties typically belong.

\citet{artetxe2020cross} introduce zero-shot cross-lingual transfer by mapping monolingual representations between languages. They also propose adding Gaussian noise to the embeddings during the fine-tuning step. 
\citet{Huang-etal-2021-improving-zero} also operate in the embedding space by constructing robust regions in the embedding space to tolerate noise in the contextual embedding.
These are not ideal strategies for closely related languages because words with similar surface forms could still be far from each other in an embedding space.

\paragraph{Surface-level noise} such as character substitutions, insertions, and deletions has been proposed as an effective data augmentation strategy for machine translation~\citep{sperber2017toward,heigold-etal-2018-robust,belinkov2018synthetic,karpukhin-etal-2019-training,vaibhav-etal-2019-improving, anastasopoulos-etal-2019-neural}. Authors report improvements in system accuracy due to more robustness towards speech recognition errors, spelling mistakes, and other naturally occurring noise in text data. 
Even though cross-lingual transfer between closely related languages has received some attention~\cite{muller2020can, sakaguchi2017robsut, zeman-etal-2017-conll, zeman-etal-2018-conll}, it has not been investigated whether this transfer can be improved with character-level noise inserted at training time.
We tackle this in our work by adding random character-level noise to the training data of a standard language and applying the model to closely related languages.

\paragraph{Exploiting orthographic similarity} to improve cross-lingual transfer between closely related languages is currently an understudied area. Relevant previous work has been done by~\citet{sharoff-2018-language}, who used orthographic similarity to refine bilingual dictionary induction. 

\paragraph{Transliteration} is another line of related work that focuses on improving the transfer between closely related languages with different alphabets~\cite{durrani-etal-2014-integrating, lin-etal-2016-leveraging, murikinati-etal-2020-transliteration, han-eisenstein-2019-unsupervised}. On the other hand, our work focuses on languages using the same script. The recent report by~\citet{muller-etal-2021-unseen} investigating transfer between the same and different alphabets involves a zero-shot task transfer which is, however, preceded by a language model training on (unlabeled) target language data. To the best of our knowledge, we are the first to focus on zero-shot transfer learning techniques for closely related languages.

\section{Method}

\begin{figure}
    \centering
    \includegraphics[width=0.48\textwidth]{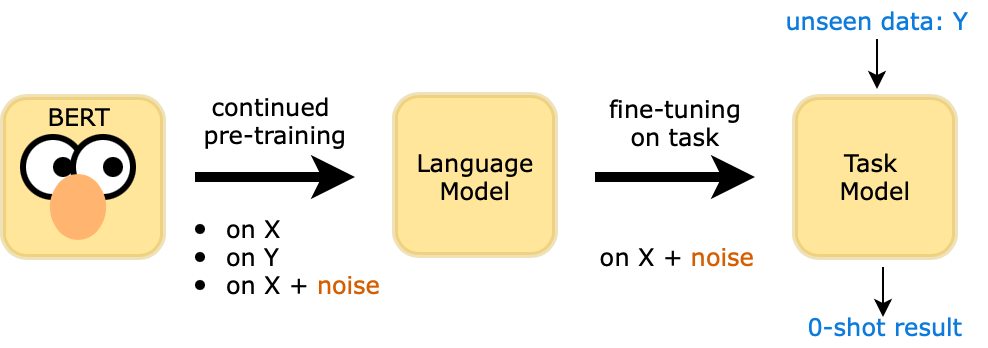}
    \caption{Methodology for zero-shot cross-lingual transfer: We first continue the pre-training of a language model (LM) on text. Then, we fine-tune the adapted LM to task $T$ in high-resource language $X$. We augment the training data for continued pre-training or fine-tuning with \textcolor{myred}{character-level noise} and \textcolor{myblue}{apply the model to task $T$ in a closely related low-resource language $Y$}.}
    \label{fig:method}
\end{figure}

Consider a high-resource language $X$ and a closely related low-resource language $Y$.
We perform zero-shot cross-lingual transfer by pre-training a model on unlabeled data from $X$ (and optionally $Y$ if available), then fine-tuning on task $T$ in language $X$. The resulting model is applied to task $T$ in language $Y$. This procedure is illustrated in Figure~\ref{fig:method}. Thus, our question is: How can we best make the model trained on $X$ generalize to $Y$?

Ideally, such a model would take surface similarity of words into account for generalization.
We hypothesize that in closely related languages, unknown words in the low-resource language are likely to correspond to similar known words in the high-resource language in function and meaning.\footnote{Please refer to Table \ref{appendix:errors} for examples.}
However, state-of-the-art language models represent words through subwords~\cite{sennrich-etal-2016-neural}.
This representation is sensitive to slight surface variations: 
minor changes to a string will lead to different segmentations and internal representations.

To account for this problem, we apply noise to the surface representation of words in language $X$.
We implement this through character-level noise, i.e., we randomly\footnote{We tested more linguistically motivated constraints on the character replacements but did not find it to have a big effect.} select $10\%-15\%$ of the tokens of a sentence\footnote{We relied on previous work by~\citet{vincent} where similar choices were made regarding the amount of noise.} excluding numbers and punctuation. One randomly selected character of the chosen token undergoes one of three possible operations: \textit{delete}, \textit{replace}, \textit{insert}, each with equal probability. 
The latter two operations work with an additional randomly selected character of the (extended) Latin Alphabet for the source language. The following sentence \textit{This is a short example.} will end up with some sort of a ``typo'', e.g., as \textit{This i\textbf{u}s a short example.}
Noise can be applied during pre-training on $X$, and/or during fine-tuning on a task $T$ for language $X$.

Another possibility to alleviate the subword representation problem is BPE-dropout~\cite{provilkov-etal-2020-bpe}, which applies different segmentations to words in a randomized fashion. BPE-dropout was originally motivated to increase robustness for morphological variance. We hypothesize that it is similarly effective for orthographic variance; see Table \ref{appendix:errors} for examples. 
Overall, both character-level noise and BPE-dropout encourage the model to learn generalizations across similar surface strings via shared subwords.

A second motivation for character-level noise is that we aim to imbue the model with different inductive biases.
For example, a model for POS tagging might learn that only a small set of words can map to closed word classes such as articles, whereas unknown words are likely to belong to an open word class such as named entities. 
Training with character-level noise will reduce this bias.

\section{Experiments}

\begin{table}
\centering
\resizebox{\columnwidth}{!}{
\begin{tabular}{c|c|c|c|c}
\textbf{} & \textbf{DE-BERT} & \textbf{DE-BERT} & \textbf{DE-BERT} & \textbf{DE-BERT} \\
\textbf{Noise} & \textbf{} & \textbf{+ GSW} & \textbf{+ DE} & \textbf{+ DE + Noise} \\

 \midrule
 \redcross  & 50.66                 & \textbf{72.1\phantom{0}}     & 52.08            & 53.88  \\
 \bluecheck  & 72.77                 & \textbf{82.11}    & 71.13            & 70.45  \\

\end{tabular}
}
\caption{POS tag accuracy for Swiss German (GSW) on different language models fine-tuned on German (DE) training data with and without noise.}
\label{tab:gsw_pos}
\end{table}

We design our experimental procedure\footnote{We work with code bases by~\citet{wolf2020huggingfaces} and~\citet{wang-etal-2021-multi-view}, multilingual BERT (mBERT), and the data sets' default splits. Most of the corpora we work with were provided by the Universal Dependency project (UD,~\citet{nivre-etal-2016-universal}); refer to Appendix \ref{appendix:UD} for details.} to answer the following question: Does character-level noise improve zero-shot transfer to closely related languages? Within three controlled experiments, we ablate the importance of the noise-augmentation strategy. We select two cross-lingual tasks: 1) POS tagging ($15\%$ noise) and 2) topic classification ($10\%$ noise). 
While the former task illustrates the strategy's potential on word level, the latter provides insight into how much it helps on text level. 

\subsection{POS Tagging for Swiss German Dialects}

As base models, we use the German ``dbmdz'' BERT\footnote{\url{https://github.com/dbmdz/berts\#german-bert}} (DE-BERT) and mBERT~\cite{devlin-etal-2019-bert}. 
We continue pre-training on the \textit{SwissCrawl} corpus~\cite{linder2020crawler} for the Swiss German (GSW) LM-adaptation and the DE part of \textit{The Credit Suisse News Corpus} 
\cite{CSNewsCorpusReleasev052019} for the German LM-adaptation. For task fine-tuning, we use a DE UD treebank and for the evaluation a part of \textit{NOAH's Corpus}~\cite{hollenstein-aepli-2014-compilation}.

As shown in Table~\ref{tab:gsw_pos}, all settings profit from fine-tuning with noise and bring about improvements of up to $22$ percentage points (DE-BERT without pre-training).
The best result with an accuracy of $82.11\%$ for zero-shot GSW POS tagging is achieved with a GSW-adapted language model and task fine-tuning on a noised DE corpus. Considering the case where no GSW text data is available for language model adaptation, we still achieve an accuracy of $77.11\%$ for zero-shot GSW POS tagging with mBERT fine-tuned on noised DE data (see Table~\ref{tab:mbert}). 

\subsection{POS Tagging with mBERT}

We fine-tune mBERT on a UD corpus of a language already seen during mBERT pre-training: \textcolor{myblue}{DE}, Finnish (\textcolor{mygreen}{FI}), Swedish (\textcolor{myred}{SV}), French (\textcolor{myorange}{FR}), or Icelandic (\textcolor{myred}{IS},~\citet{icelandic}) and test on a closely related language variety absent from mBERT: \textcolor{myblue}{GSW}, Old French (\textcolor{myorange}{OFR}), Livvi (\textcolor{mygreen}{OLO},~\citet{pirinen-2019-building}), Karelian (\textcolor{mygreen}{KRL},~\citet{pirinen-2019-building}), or Faroese (\textcolor{myred}{FO},~\citet{tyers-etal-2018-multi}). In addition to noise, we added experiments with a BPE-dropout of 0.1 (empirically selected) during the fine-tuning step.

Table~\ref{tab:mbert} illustrates that the method works well for closely related language varieties (upper part) but less for other language pairs, which are more distant (lower part). We do see an occasional improvement for more distant language pairs, but they are generally smaller and less consistent than the improvements for the closely related languages we evaluated. 

Furthermore, we have to consider the (much) lower baseline where an accuracy gain does not have the same impact.
Hence, the two strategies BPE-dropout and noise improve the zero-shot performance for POS tagging over several closely related language pairs. 
While BPE-dropout shows some performance gain over the baseline, character-level noise adds additional accuracy points.

\begin{table}
\resizebox{\columnwidth}{!}{
\begin{tabular}{c|cccc}
\textbf{\small{}}        & \textbf{\small{}} & \textbf{\small{BPE-}} & \textbf{\small{}} & \textbf{\small{BPE-Drop-}}   \\ 
\textbf{\small{Languages}} & \textbf{\small{Baseline}} & \textbf{\small{Dropout}} & \textbf{\small{Noise}} & \textbf{\small{-out+Noise}} \\  
\midrule
\textcolor{myblue}{DE}$\rightarrow$\textcolor{myblue}{GSW} & 73.14       & 76.48       & 77.11          & \textbf{78.13}                    \\
\textcolor{mygreen}{FI}$\rightarrow$\textcolor{mygreen}{OLO} & 69.32       & 69.66       & \textbf{73.03} & 71.76                  \\
\textcolor{mygreen}{FI}$\rightarrow$\textcolor{mygreen}{KRL} & 72.44       & 76.35       & \textbf{79.18} & 78.57                    \\
\textcolor{myred}{SV}$\rightarrow$\textcolor{myred}{FO}  & 84.76       & 86.20       & \textbf{87.63} & 87.31                    \\
\textcolor{myred}{IS}$\rightarrow$\textcolor{myred}{FO}  & 85.94       & 86.80            & 87.43          & \textbf{87.46}                             \\
\textcolor{myorange}{FR}$\rightarrow$\textcolor{myorange}{OFR} & 63.42       & 66.65       & 66.73          & \textbf{67.27}      \\ \midrule  
\textcolor{myblue}{DE}$\rightarrow$\textcolor{myred}{FO} & 81.74 & 81.34& 81.38 & \bf{82.27}\\ 
\textcolor{myblue}{DE}$\rightarrow$\textcolor{mygreen}{OLO} & \bf{52.63} & 52.09 & 51.10 & 49.26 \\  
\textcolor{myblue}{DE}$\rightarrow$\textcolor{mygreen}{KRL} & \bf{57.51} & 57.47 & 55.71 & 53.37 \\
\textcolor{myblue}{DE}$\rightarrow$\textcolor{myorange}{OFR} & \bf{44.08} & 39.17 & 38.32 & 40.03 \\
\textcolor{myorange}{FR}$\rightarrow$\textcolor{mygreen}{OLO} & 56.49 & 56.72 & \bf{58.59} & 56.64\\
\textcolor{myorange}{FR}$\rightarrow$\textcolor{mygreen}{KRL} & 59.46 & 62.27 & \bf{64.52} & 64.15\\
\textcolor{myorange}{FR}$\rightarrow$\textcolor{myred}{FO} & 81.13 & 82.09 & 81.81 & \bf{82.62} \\  
\end{tabular}
}

\caption{Zero-shot POS tagging accuracy of different strategies for several languages (TRAIN$\rightarrow$TEST).
The training and test languages are closely related in the upper but not in the lower part of the table as indicated by the colors (\textcolor{mygreen}{Finnic}, \textcolor{myblue}{West Germanic}, \textcolor{myred}{North Germanic}, and \textcolor{myorange}{Western Romance} language branches.) Noise consistently adds additional accuracy points beyond BPE-dropout performance increase.}
\label{tab:mbert}
\end{table}

\subsection{Cross-dialect Topic Identification}

We work with mBERT and \textit{MOROCO: The Moldavian and Romanian Dialectal Corpus}~\cite{butnaru2019moroco}. The data set contains 33,564 news domain text samples, each belonging to one of six topics (culture, finance, politics, science, sports, tech). 
We fine-tune mBERT on topic identification on Moldavian (MD) and evaluate on Romanian (RO) and vice versa.
We emphasize the difference between this sentence-level task and the previous word-level task. While POS tagging works on the word form and can benefit from transferring prior information about probable POS sequences, topic classification is mainly meaning-oriented, making a transfer more challenging.

\begin{table}
\centering
\scalebox{0.7}{
\begin{tabular}{c|c|c|c}
\textbf{Training}   & \textbf{Noise} & \textbf{Test}       & \textbf{Accuracy}     \\ \midrule 
\multirow{2}{*}{MD} & \redcross   & \multirow{2}{*}{RO} & 63.34     \\
                    & \bluecheck   &                     & \textbf{68.48}                            \\ \midrule
\multirow{2}{*}{RO} & \redcross    & \multirow{2}{*}{MD} & 81.65    \\
                    & \bluecheck   &                     & \textbf{83.01}                           \\
\end{tabular}
}
\caption{Results for Moldavian (MD) vs.~Romanian (RO) cross-dialect topic identification. Training with noise improves the transfer by $5.1$ (MD$\to$RO) respectively $1.3$ (RO$\to$MD) percentage points.}
\label{tab:mrc}
\end{table}

Topic identification results in Table \ref{tab:mrc} show that fine-tuning with noise consistently improves the accuracy.
However, noise-augmented training data appears to have a more substantial effect when transferring from MD to RO ($5.1$ percentage points) than vice-versa ($1.4$ percentage points).
This is interesting given that RO represents the high-resource standard language in this context (being one of the languages used to train mBERT), while MD is its low-resource variety.
We conjecture that this is caused by the fact that the model trained in MD struggles with word meaning and is, therefore, more sensitive to variations than its RO counterpart.

\section{Analysis}

\begin{figure}
    \centering
    \includegraphics[width=0.48\textwidth]{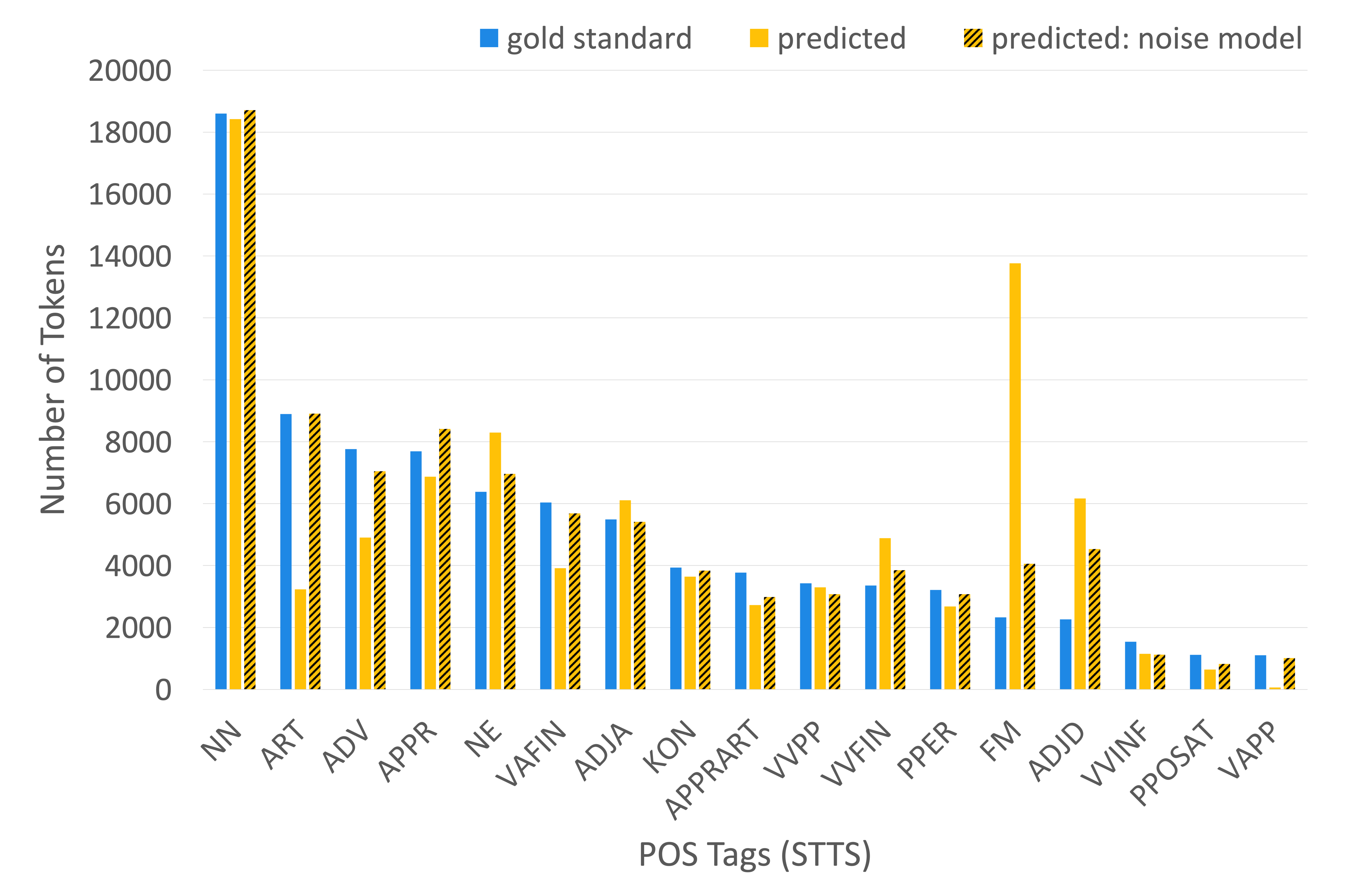}
    \caption{Number of tokens per POS tag in the gold standard vs.~predictions of two models, fine-tuned with and without noise. Only the most frequent STTS tags are displayed.}
    \label{fig:tag_stats}
\end{figure}

Figure \ref{fig:tag_stats} illustrates the prediction differences of a model trained with and without noise. 
We observe that the models trained without noise have learned a tendency towards labeling unknown words as open-class words such as names (\texttt{NE}) or adjectives (\texttt{ADJD}), with the label for foreign words (\texttt{FM}) being massively overpredicted, while it tends to under-generate closed-class tags such as articles (\texttt{ART}) or adverbs (\texttt{ADV}).
In contrast, the model trained with noise comes much closer to the gold standard tag distribution. It has learned to rely more on probable POS tag sequences than on the surface form of a token. 
Consider e.g. the GSW article \textit{d} (DE \textit{die}; the). In the DE training corpus, the token appears only as a foreign word (\texttt{FM}) because it also happens to be a French word, but the model trained with noise is more likely to correctly tag it as an article, relying more on context than just strict mappings.\footnote{For more examples, please refer to the Appendix \ref{appendix:analysis}.}

\begin{figure}
    \centering
    \includegraphics[width=0.48\textwidth]{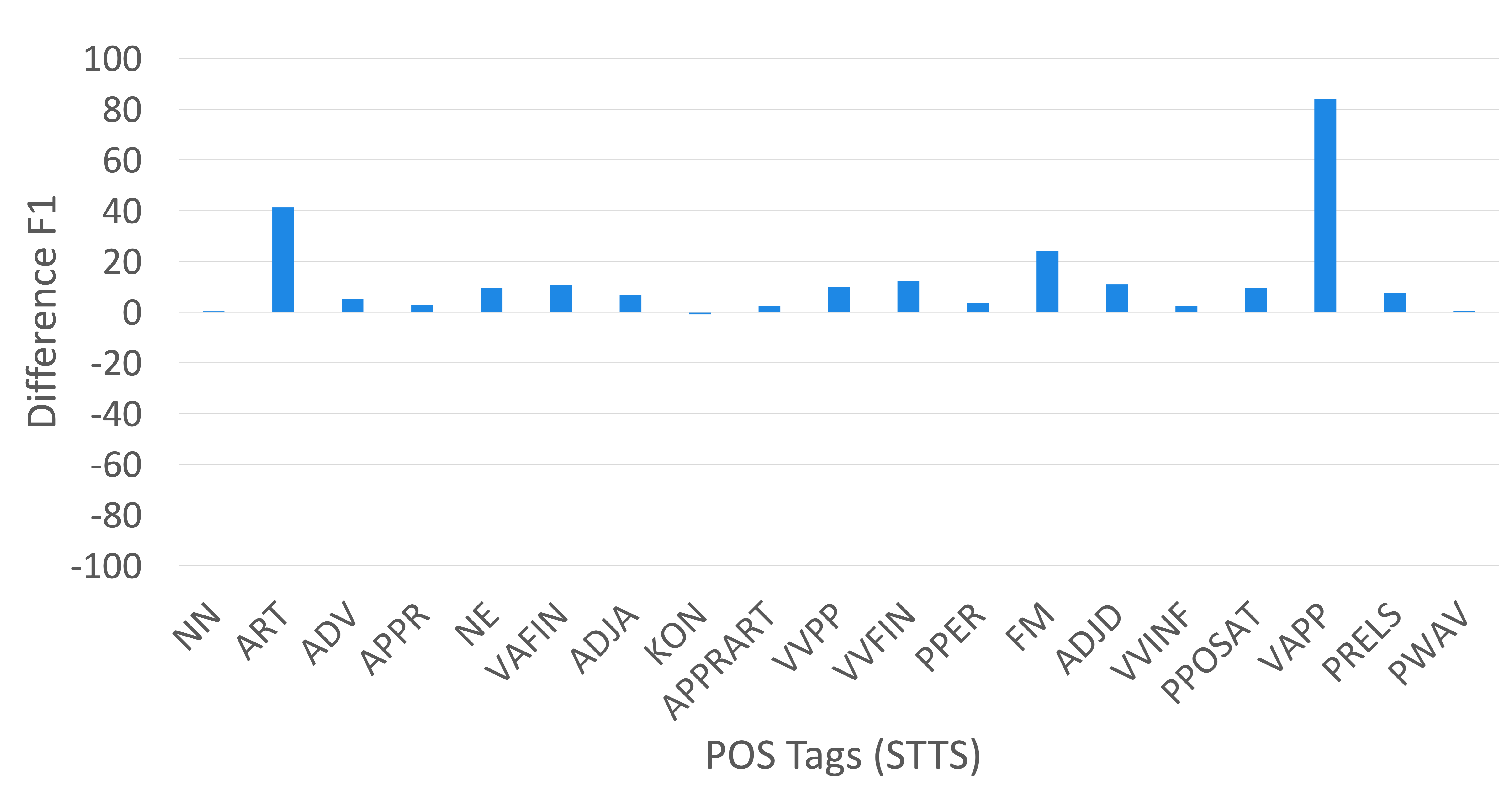}
    \caption{Per-type F1 change of the most frequent STTS tags illustrating which tags profit the most when the model is fine-tuned with noise.}
    \label{fig:f1_diff}
\end{figure}

Figure \ref{fig:f1_diff} depicts the per-type F1 change for the most frequent STTS~\cite{STTS} tags. The past participles of the auxiliary verbs (\texttt{VAPP}) form another closed class which profits substantially from a model focusing on tag sequences given the compound structure of the perfect tense. As Swiss German does not have a simple past, the perfect tense is much more frequent than it is in German.

\subsection{Conditional Random Field}

Given that one potential consequence of adding noise is that the model relies more on surrounding context and probable POS tag sequences (rather than strict word-level mappings), we compare our results to a method that explicitly models tag sequences, Bi-LSTMs+CRF (bidirectional long-short-term memory + conditional random field). This method was used to achieve state-of-the-art performance for POS tagging~\cite{huang2015bidirectional}. For the implementation, we added a CRF layer on top of BERT.\footnote{Making use of the TorchCRF library: \url{https://github.com/s14t284/TorchCRF}.}

\begin{table}[ht]
\centering
\scalebox{0.7}{
\begin{tabular}{c|c|c}
\textbf{Noise}      & \textbf{mBERT} & \textbf{mBERT+CRF}  \\ \midrule
\textbf{\redcross }         & 70.24          & 69.26          \\
\textbf{\bluecheck}        & \textbf{78.57} & 76.90     
\end{tabular}
}
\caption{Zero-shot POS tag accuracy for Swiss German on mBERT \& mBERT+CRF models trained on German with and without noise.}
\label{tab:crf}
\end{table}

The added CRF layer did not improve the performance of the fine-tuned 
 mBERT model for zero-shot POS tagging for Swiss German trained on German, as presented in Table \ref{tab:crf}. In contrast, noise injection has proven effective in both configurations.

\section{Discussion}

Our investigation into cross-lingual zero-shot transfer between closely related languages demonstrates that simple data augmentation with character-level noise can successfully improve transfer, with absolute improvements ranging from $1.4$ (RO$\to$MO transfer of topic identification) to $22$ (DE$\to$GSW transfer of POS tagging in the case of DE-BERT without pre-training) percentage points.

The examination of prediction errors shows that a baseline BERT model has learned heuristics for unseen words that are undesirable for transfer between closely related languages. In contrast, a model trained with noise can combat this bias without substantial performance losses in the source languages.

We expect that the final effectiveness of using character-level noise for zero-shot cross-lingual transfer depends on the task and language characteristics.
We plan to evaluate the effect of character-level noise in a broader range of settings in future work. 
More broadly, we encourage further research that exploits surface-level word similarity for cross-lingual transfer between related languages and dialects, rather than focusing purely on vector space representations.

\section*{Acknowledgments}

We thank Yves Scherrer for input regarding data sets and the anonymous reviewers for their valuable comments. This project has received funding from the Swiss National Science Foundation (project nos.\ 191934 \& 176727).

\section*{Ethical Considerations}

Our work did not involve any new data collection or annotation and did not require in-house workers or introduce any new models and related risks. Instead, we examine how character-level noise can help the transfer between closely related languages, especially in low-resource zero-shot settings.

\bibliography{anthology,custom}
\bibliographystyle{acl_natbib}

\clearpage
\appendix

\setcounter{table}{0}
\renewcommand{\thetable}{A\arabic{table}}

\section{Appendix}

\subsection{Analysis}\label{appendix:analysis}

Table \ref{appendix:errors} shows Swiss German (GSW) words (and their corresponding standard German (DE) form and English (EN) translation) that had the highest accuracy increase when using a part-of-speech (POS) tagging model trained with character-level noise compared to the model trained without noise. 
These words were wrongly tagged with open-class tags by the baseline model. However, the model trained with noise was able to reduce this bias and thus correctly tag them with their closed-class tag.

Furthermore, in most cases, one \textit{substitution/insertion/deletion}-operation on the DE word would not suffice to get an exact match with the GSW word. This indicates that it is unnecessary to design a noise function that closely mirrors the linguistic differences between variants.

\begin{table*}[ht]
\centering
\begin{tabular}{l|llllll}
\textbf{} & \textbf{} & \textbf{} & \textbf{Most frequent} & \textbf{} & \textbf{Error reduction with} \\
\textbf{GSW} & \textbf{DE} & \textbf{EN}  & \textbf{POS without noise} & \textbf{Correct POS} & \textbf{noise (relative/absolute)} \\
\hline
ond          & und & and         & FM                                       & KON                                   & 99.00\% (-104)           \\
worde        & geworden & become & VVPP, ADJD                               & VAPP                                  & 98.73\% (-156)           \\
dr           & der & the         & FM, NE, NN                               & ART                                   & 98.46\% (-128)           \\
häd          & hat & had         & VVFIN, FM                                & VAFIN                                 & 98.21\% (-55)            \\
gsi          & gewesen & been    & VVPP, FM                                 & VAPP                                  & 98.19\% (-434)           \\
vu           & von & from        & FM, NE                                   & APPR                                  & 98.18\% (-108)           \\
eme          & einem & a         & ADJA, FM                                 & ART                                   & 97.96\% (-48)            \\
grad         & gerade & just     & ADJD                                     & ADV                                   & 97.59\% (-81)            \\
vum          & vom & from the    & FM, APPR                                 & APPRART                               & 96.49\% (-55)            \\
de           & der & the         & FM, NE, ADJA                             & ART                                   & 95.76\% (-1558)         
\end{tabular}
\caption{Swiss German (GSW) words (and their corresponding standard German (DE) form and English (EN) translation) with the highest error reduction using a part-of-speech (POS) tagging model trained with character-level noise compared to the model trained without noise.}
\label{appendix:errors}
\end{table*}

\subsection{Data Sets}\label{appendix:UD}

\subsubsection{Universal Dependencies}

Table \ref{appendix:tabud} contains the Universal Dependencies treebanks (UD,~\citet{nivre-etal-2016-universal}) we used in this work. The treebanks can be downloaded via \url{https://universaldependencies.org/#download}.

\begin{table*}
\centering
\begin{tabular}{l|llll}
\bf{Usage} & \bf{Language (ISO)} & \bf{Language Branch} & \bf{Treebank} &  \bf{\# Sentences} \\ \midrule
\multirow{5}{*}{\textbf{Training}}  & Finnish (FI) & Finnic & TDT & 15K \\
& French (FR) & Western Romance & GSD &  16K\\
& German (DE) & West Germanic & HDT & 190K \\
& Icelandic (IS) & North Germanic & IcePaHC & 39K \\
& Swedish (SV) & North Germanic &  Talbanken & 5K \\ \midrule
\multirow{4}{*}{\textbf{Test}} & Faroese (FAO) & North Germanic & OFT & 1208 \\
& Karelian (KRL) & Finnic & KKPP &  228 \\
& Livvi (OLO) & Finnic &  KKPP &  125  \\
& Old French (OFR) & Western Romance & SRCMF &  1927 \\
\end{tabular}

\caption{Universal Dependency treebanks we used for our experiments with the number of sentences ("\# Sentences") we used for training or testing (specified in "usage").} 
\label{appendix:tabud}
\end{table*}

\subsubsection{Other Data Sets}

Table \ref{appendix:tabothers} shows data sets apart from UD that we used in this work.

\begin{table*}
\centering
\resizebox{\textwidth}{!}{
\begin{tabular}{l|ll}
\bf{Corpus name (Language)} & \bf{Size} & \bf{Link}\\ \midrule
Moroco (MD \& RO) & 33.5K text samples & \url{https://github.com/butnaruandrei/MOROCO} \\
NOAH's Corpus (GSW) & 7.3K sentences & \url{https://noe-eva.github.io/NOAH-Corpus/} \\
SwissCrawl (GSW) & 500K sentences &  \url{https://icosys.ch/swisscrawl} \\
The Credit Suisse News Corpus (DE) & 105K sentences & \url{https://pub.cl.uzh.ch/projects/b4c/en/corpora.php} \\
\end{tabular}
}
\caption{Data sets we used for our experiments in addition to the UD treebanks in Table \ref{appendix:tabud}.}
\label{appendix:tabothers}
\end{table*}

\end{document}